\documentclass[conference]{IEEEtran}
\IEEEoverridecommandlockouts
\usepackage{booktabs}
\usepackage{comment}
\usepackage{cite}
\usepackage{amsmath,amssymb,amsfonts}
\usepackage{algorithmic}
\usepackage{graphicx}
\usepackage{textcomp}
\usepackage{xcolor}
\usepackage{float}
\usepackage{afterpage}
\def\BibTeX{{\rm B\kern-.05em{\sc i\kern-.025em b}\kern-.08em
    T\kern-.1667em\lower.7ex\hbox{E}\kern-.125emX}}
\begin{document}

\title{Normalization through Fine-tuning: Understanding Wav2vec 2.0 Embeddings for Phonetic Analysis\\
{}
}

\author{
\IEEEauthorblockN{Yiming Wang, Yi Yang}
\IEEEauthorblockA{\textit{School of the Gifted Young} \\
\textit{University of Science and Technology of China}\\
Hefei, China \\
Emails: wangyiming@mail.ustc.edu.cn, yanggnay@mail.ustc.edu.cn}
\and
\IEEEauthorblockN{Jiahong Yuan}
\IEEEauthorblockA{\textit{School of Humanities and Social Sciences} \\
\textit{University of Science and Technology of China}\\
Hefei, China \\
Email: jiahongyuan@ustc.edu.cn}
}

\maketitle

\begin{abstract}
Phonetic normalization plays a crucial role in speech recognition and analysis, ensuring the comparability of features derived from raw audio data. However, in the current paradigm of fine-tuning pre-trained large transformer models, phonetic normalization is not deemed a necessary step; instead, it is implicitly executed within the models. This study investigates the normalization process within transformer models, especially wav2vec 2.0. Through a comprehensive analysis of embeddings from models fine-tuned for various tasks, our results demonstrate that fine-tuning wav2vec 2.0 effectively achieves phonetic normalization by selectively suppressing task-irrelevant information. We found that models fine-tuned for multiple tasks retain information for both tasks without compromising performance, and that suppressing task-irrelevant information is not necessary for effective classification. These findings provide new insights into how phonetic normalization can be flexibly achieved in speech models and how it is realized in human speech perception.

\end{abstract}







\begin{IEEEkeywords}
speech representation, normalization, wav2vec 2.0, Transformer
\end{IEEEkeywords}

\section{Introduction}

Transformer-based speech models, such as wav2vec 2.0, have shown remarkable success in phonetics-related tasks, including phoneme recognition, tone recognition, and speaker identification. To leverage wav2vec 2.0 embeddings for phonetic analysis, it is essential to understand how these embeddings encode information, particularly with regard to phonetic normalization. For example, when analyzing the acoustic differences among lexical tones to uncover inherent tonal features in a language, speaker-specific variations, e.g., females have a higher pitch than males, must be normalized to allow for meaningful comparisons of tonal features across speakers.

Probing experiments have demonstrated that the embeddings of pre-trained wav2vec 2.0 models encode various types of information, with different layers capturing different aspects. Our study takes this a step further by investigating whether and how fine-tuning wav2vec 2.0 models for a specific task can suppress information irrelevant to that task. Moreover, we examine how the embeddings differ when the model is fine-tuned for multiple tasks compared to a single task. 

Our main findings are summarized as follows:
\begin{itemize}
\item {Phonetic normalization is achieved through the fine-tuning of wav2vec 2.0. The process, from the initial to the final layers of a fine-tuned Transformer, varies depending on the type of information: for tones and finals, the information is initially enhanced and subsequently suppressed, while for speaker sex, the information is progressively suppressed throught the layers. The observed differences can likely be attributed to the way information is encoded across the various layers of the pretrained wav2vec 2.0 model.}

\item {The embeddings from a model fine-tuned for multiple tasks, such as tone and sex classification, effectively retain information of both tone and sex. The effectiveness of these embeddings is comparable to those fine-tuned for each task separately.}

\item {All task-irrelevant information is suppressed in fine-tuned models. For instance, the embeddings of a model fine-tined for tone recognition suppresses both sex and final information. This is not an obvious outcome from a machine learning perspective, as embeddings containing both tone and sex information perform equally well for optimizing the objective of tone classification. In other words, when fine-tuning for tone classification, suppressing sex information is not strictly necessary.}

\item {The combination of results 2 and 3 have an important implication for phonetic analysis. Phonetic normalization does not necessarily require the suppression of task-irrelevant information. For example, sex or final information can be either suppressed or preserved in features that effectively distinguish tones. This finding provides new insights into how phonetic normalization can be flexibly achieved in speech models and how it is realized in human speech perception.}

\end{itemize}

\section{Related work}
\subsection{Normalization}
Normalization in human speech perception has long been studied \cite{Johnson&Sjerps-2021}. \cite{Miller-1953} investigated the effect of F0 on the perception of vowels using synthetic sounds, showing that the perceptual category boundary between vowels shifted when the F0 was doubled. \cite{Honorof&Whalen-2005} found that listeners can determine where the pitch of an isolated vowel, produced by a stranger, falls within that stranger's pitch range, without prior familiarity or context. \cite{Tang&Hamilton&Chang-2017} used high-density electrocorticography to record neural population activity while participants listened to sentences that varied in intonational pitch contour, phonetic content, and speaker. They found that the cortical representation of intonation relies on relative-pitch encoding, rather than absolute-pitch encoding.

In phonetic research, numerous vowel normalization methods have been proposed in the literature \cite{Adank-2004,Voeten-2022}. Regarding pitch and tone normalization, commonly utilized techniques include z-scores, semitones, and the 5-point scale \cite{Rose-1987,Zhang-2018}. In speech technology, techniques such as speaker adaptive training (SAT), vocal tract length normalization (VTLN), and Maximum Likelihood Linear Regression (MLLR) are extensively employed for speaker normalization and adaptation \cite{Giuliani-2006}.

In deep learning, a variety of normalization techniques have been developed to speed up training and mitigate issues related to vanishing and exploding gradients. These include batch normalization, layer normalization, weight normalization, and spectral normalization, among others. It's important to note that these methods do not incorporate explicit speaker or phonetic normalization as discussed above.
\subsection{wav2vec 2.0}
wav2vec 2.0 is a self-supervised learning framework based on Transformers that can learn speech representations from raw audio data \cite{baevski2020wav2vec}. The framework processes the speech signal with a multilayer convolutional network to extract latent features, which are then fed into vector quantization and transformer networks. It is pre-trained on a noise contrastive binary classification task, which enables the model to capture a rich amount of information about speech, as demonstrated by probing experiments that show their effectiveness on a wide range of tasks \cite{shah2021probing,ma2021probing,Bartelds-2022}.%

The pre-trained models of wav2vec2 can be fine-tuned for speech recognition using labeled data \cite{baevski2020wav2vec}. \cite{yuan2021suprasegmental} fine-tuned wav2vec 2.0 with a CTC loss to recognize suprasegmentals such as syllables, tones, and pitch accents. Their results demonstrated a 50\% error reduction in Mandarin tone recognition compared to previous studies. \cite{yuan-2023} proposed fine-tuning wav2vec 2.0 with a cross-entropy loss to classify tones in an utterance on a frame-by-frame basis, which not only improves tone
classification accuracy but also generates frame-level representations suitable for tonal analysis.

Using features extracted from a pre-trained wav2vec 2.0 model for phoneme recognition, \cite{Baevski-2021} discovered that the features from the first 10 layers as well as the last few layers provide very poor performance, while layers 15-19 achieve the best phoneme recognition accuracy. In this study, we will investigate the performance of features extracted from various layers of fine-tuned wav2vec 2.0 models across multiple tasks.
\section{Data and method}
\subsection{Data}
Our experiments were conducted on the Aishell-1 dataset \cite{Bu2017AISHELL1AO}, which is widely used for Mandarin ASR. It consists of 165 hours of read speech and word transcriptions in Mandarin Chinese from 400 speakers. The dataset is divided into train, dev, and test sets with 150, 10, and 5 hours of speech, respectively.

\subsection{Forced Alignment}
We trained an HMM-GMM based forced aligner on the dataset using the HTK toolkit\footnote{https://htk.eng.cam.ac.uk/} and the pronouncing dictionary included in the dataset. The dictionary transcribes words into initials and tonal finals in Pinyin, a Roman alphabet system of phonetic transcription for Mandarin Chinese.

From forced alignment, we assigned a label to each frame in an utterance. The central frame within the boundary of a tone or phone received the corresponding label whereas all other frames were labeled as 'O', a special category that will be ignored in computing cross-entropy loss in fine-tuning. The framerate, set at 20ms, matches the output framerate of wav2vec 2.0. 

\subsection{Fine-tuning wav2vec 2.0 for framewise classification}
Pairs of audio and frame-aligned labels in the training set were used to fine-tune a wav2vec 2.0 large model pre-trained on 960 hours of Librispeech audio (libri960\_big.pt) for framewise classification. 
In this task, every frame in an utterance is classified into a label. In a conventional approach to framewise classification (e.g. \cite{Ryant-2014}), the features of a frame are locally extracted and the frames are largely independent of each other. In our approach, features of all frames in an utterance are interdependent due to the self-attention mechanism in the transformer model.

During training, a randomly initialized linear projection is added on top of the contextual representations of wav2vec 2.0 to map the representations into label tokens. The labels belong to a set of $C$ categories, excluding a specific category 'O', which is ignored in the loss computation. The network's predictions $\hat{y}_{i,c}$ represent the probability that token $i$ belongs to category $c$. The true labels are represented by a one-hot encoded vector $y_{i,c}$, where $y_{i,c}=1$ indicates that token $i$ belongs to category $c$, and $y_{i,c}=0$ otherwise. 

The entire network is then optimized by minimizing a cross-entropy loss as described in Equation~\ref{eq:loss}, with the category of 'O' being excluded during the loss calculation. Specifically, for each token $i$ and each category $c \in \{1, 2, \dots, C\}, c \ne O$, the cross-entropy loss is computed as:

\begin{equation}
    \label{eq:loss}
    \text{Cross-Entropy}(y, \hat{y}) = -\frac{1}{N}\sum_{i=1}^{N} \sum_{c=1,c \ne O}^{C} y_{i,c} \log(\hat{y}_{i,c})
\end{equation}

where $N$ is the number of tokens in the dataset, $y_{i,c}$ is the true label for token $i$, and $\hat{y}_{i,c}$ is the model's predicted probability for token $i$ being in category $c$. The experiment was carried out on the Fairseq platform\footnote{https://github.com/facebookresearch/fairseq}.





Initially, only the output classifier is trained for the first 10k updates, after which the Transformer is updated as well. We set the max tokens to 1 million, which is equivalent to 62.5 seconds of audio with a sampling rate of 16 kHz, and the learning rate to 1e-5. The total number of updates is set to 100k. 

During inference, each audio utterance was fed forward through the fine-tuned model. The model output at the central frame of every phone or tone in the test set was compared with the reference label to calculate the classification accuracy.  The features from each of the 24 Transformer layers were extracted at the central frame of every phone or tone in the test set for analysis and visualization. These features have a dimension of 1024.

It’s worth noting that while during training only one frame of each tone or phone has a target label and all other frames are labeled as category ‘O’, which is ignored in computing loss, during inference no frames are classified as ‘O’, and in most cases, all frames within the boundary of a phone or tone get the same classification. Examples of training and prediction labels are listed below:

\begin{itemize}
\item \textit{Training labels}: O O O O O T1 O O O O O T5 O ...
\item \textit{Predictions}: T1 T1 T1 T1 T1 T1 T1 T1 T1 T5 T5 T5 T5 ...
\end{itemize}

\subsection{Multitask Training for Simultaneous Encoding of Sex, Tone, and Final}

In order to test whether the model can simultaneously encode sex, tone, and final, we employed a multitask training approach. During training, the model was designed to perform two classification tasks, denoted as task1 and task2. Each task contributes a loss term, with $L_1$ and $L_2$ representing the losses for task1 and task2, respectively. The final loss function used for optimization is given by:

\[
L = L_1 + L_2
\]

This combined loss function allows the model to optimize both tasks simultaneously. By employing this multitask learning strategy, we observed the performance of the model across both classification tasks to evaluate its ability to encode the multiple attributes of sex, tone, and final in a single unified representation.
\section{Results}
\vspace{0.5cm}
\subsection{Classification accuracy}

\begin{table}[!h]
    \centering
    \caption{Accuracy for mono-task and muti-task trained model}
    \label{table:acc_for_models}
    \begin{tabular}{lccccc}
        \toprule
        & t & st & f & sf & tf  \\
        \midrule
        Tone accuracy & 95.95\% & 95.42\% & - & - & 95.15\%  \\
        Final accuracy & - & - & 98.37\% & 98.42\% & 97.00\%  \\
        \bottomrule
    \end{tabular}
\end{table}

Using the method described above, we fine-tuned wav2vec 2.0 for tone, phone, and sex classification, respectively. To test whether different tasks can be encoded simultaneously, we also trained the combinations of the tasks which are sex\&tone(st), sex\&final(sf), and tone\&final(tf). There are five tones in total, labeled T1 to T5 (with T5 representing the neutral tone), and 39 finals. The classification accuracy on the test set is reported in Tabel \ref{table:acc_for_models}, demonstrating the excellent performance of the fine-tuned models on these tasks. A plausible explanation is that, although only a single frame of each tone or phone is utilized to train classifiers, its features encapsulate rich contextual information learned through the self-attention mechanism in Transformer. Meanwhile, the similar performance between the multi-task model and mono-task model shows that sex and tone/final can be simultaneously encoded within a shared feature space.

\subsection{Normalization}
To observe how normalization is performed in the model fine-tuned for different tasks, we extracted features from all Transformer layers of the finetuned and pre-trained models, compared these features. We reduced the features to 100 dimensions with PCA and computed correlation through Singular Vector Canonical Correlation Analysis for Deep Learning Dynamics and Interpretability (SVCCA)\cite{raghu2017svcca, morcos2018insights}, and utilized UMAP for visualization. The features were extracted at the central frame of all tones/finals in the test set.

In the SVCCA analysis, we computed the correlation between the extracted features and tones, sex, and finals, respectively. A high correlation indicates a strong similarity between the features and the corresponding labels.


\begin{figure}[!b]
  \centering
  \includegraphics[width=0.9\linewidth]{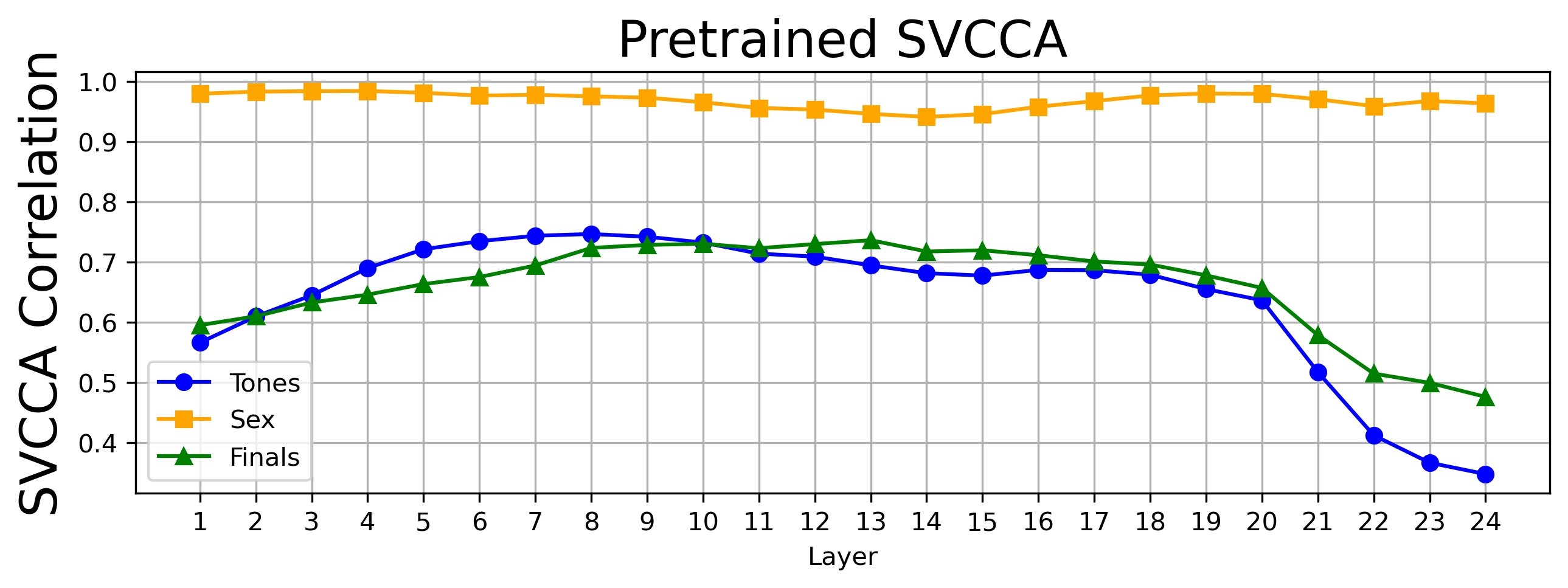}
  \caption{SVCCA correlations on features extracted from different layers of pre-trained model across three different tasks}
  \label{fig:lda_pretrained}
\end{figure}

\begin{figure}[!t]
  \centering
  \includegraphics[width=0.9\linewidth]{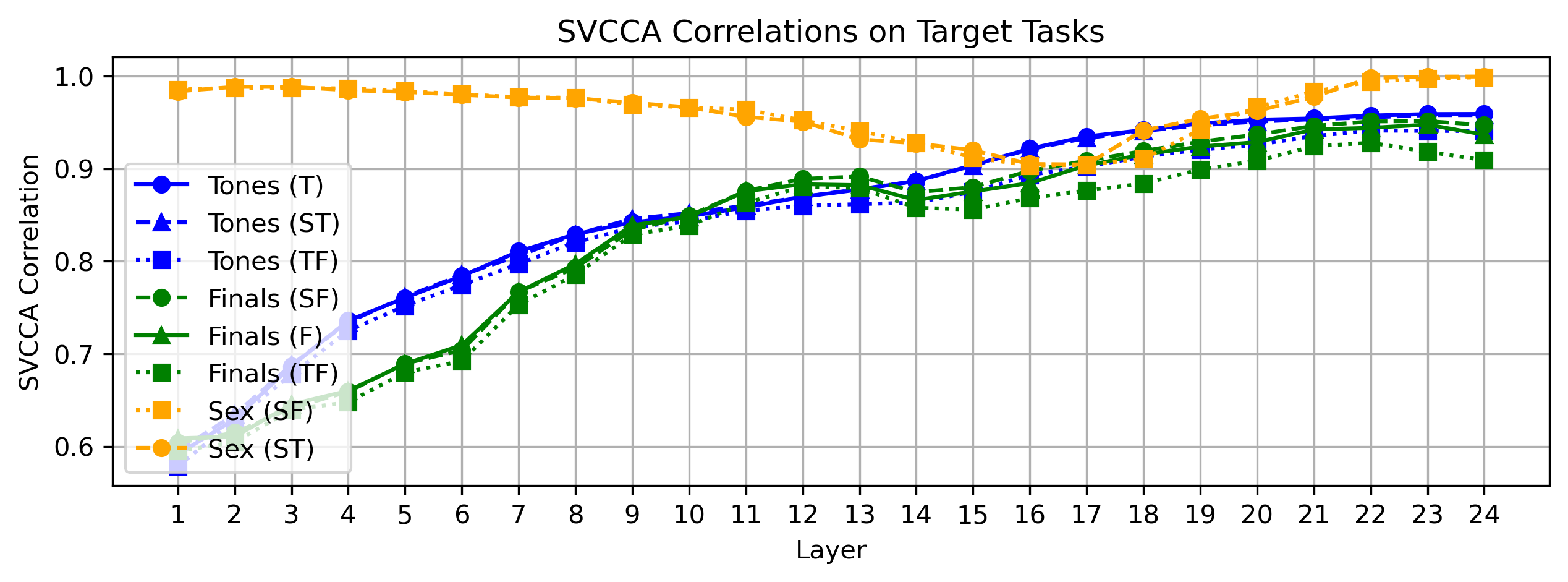}
  \caption{SVCCA correlations on features extracted from models optimized for different target tasks}
  \label{fig:target}
\end{figure}

\begin{figure}[!b]
  \centering
  \includegraphics[width=0.9\linewidth]{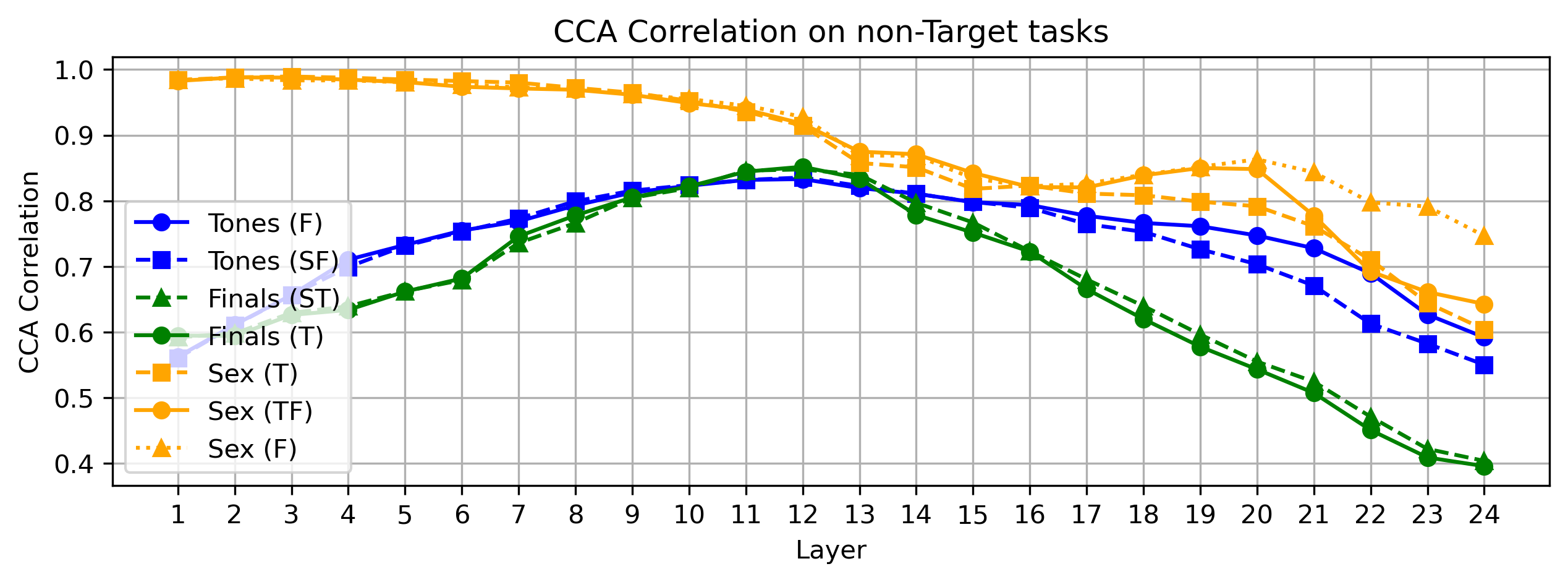}
  \caption{SVCCA correlations on features extracted from models optimized for non-target tasks}
  \label{fig:non-target}
\end{figure}
Figure~\ref{fig:lda_pretrained} shows the SVCCA correlation of features extracted from the pre-trained model at every layer of the Transformer. A couple of interesting observations can be made. Firstly, the features from the pre-trained model exhibit superior performance in sex correlation compared to tone and phone. The sex identification was conducted at the segment level rather than the utterance level. This means that the model must classify each segment of tone/final into a sex, which is much more complex than using a whole utterance to compute. Secondly, features from different layers demonstrate varying performance. Specifically, for sex correlation, the lower and upper layers are higher than the middle layers, whereas for tone and final correlation, the middle layers are higher.



Figure~\ref{fig:target} compares the normalization processes under different training objectives. For the task targeted during fine-tuning, the SVCCA correlations exhibit an upward trend from the middle layers to the final layers, indicating that this part of the network is primarily influenced by the fine-tuning task. In contrast, in the earlier layers, sex correlation remains high while tone and final correlations are lower. This pattern reflects the behavior of the initial pre-trained model, where sex characteristics dominate before the model is adjusted by fine-tuning. These observations highlight how different training tasks affect the model's layer-wise representations and the varying degrees of normalization that occur throughout the network. Additionally, Figure~\ref{fig:non-target} illustrates how the SVCCA correlations for non-target tasks change across the model layers. We observe that these correlations are gradually normalized after the middle layers, indicating that the model increasingly focuses on the target task as fine-tuning progresses, while non-target task features are progressively suppressed.

Features undergo transformations across different layers, resulting in new representations. This transformation process is observed as a decrease in the correlation of SVCCA when features are modified in ways that do not benefit downstream tasks. Conversely, if the transformations render the features useful, they are re-extracted in later layers, indicated by an increase in SVCCA correlations. Features that are irrelevant to the task at hand are normalized and discarded.

\afterpage{%
  \begin{figure}[h!]
    \centering
    \includegraphics[width=0.8\linewidth]{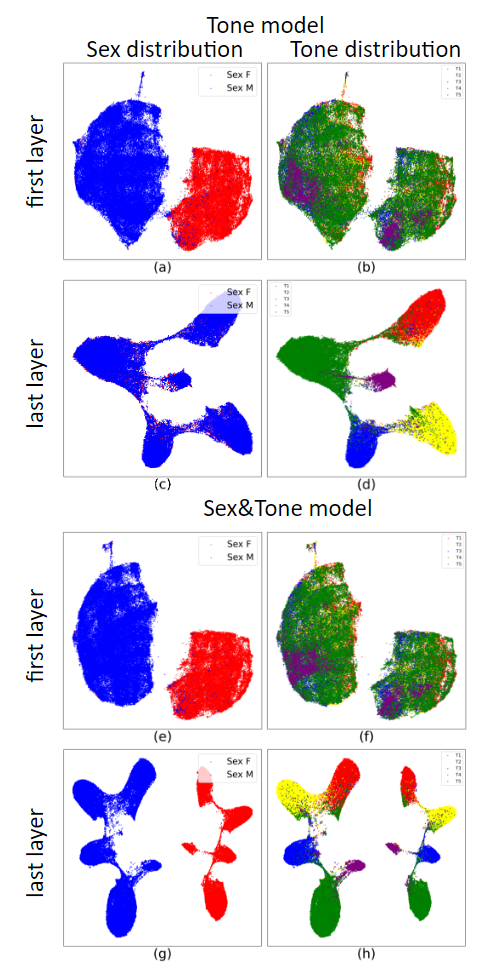}
    \caption{Classification accuracy difference between the model fine-tuned for tone classification and the pre-trained model.}
    \label{fig:transformer_features}
  \end{figure}
}

The sex normalization in the model fine-tuned for tone classification is also evident in the UMAP\footnote{https://umap-learn.readthedocs.io/en/latest/} visualization of the features. Figure~\ref{fig:transformer_features} display the features extracted from the first Transformer layer of two fine-tuned models: one trained on the tone classification task (a,b,c,d) and the other on the sex\&tone classification task (e,f,g,h). In the visualizations on the left, features are colored based on sex, while on the right, they are colored based on tone. In the first layer, both models divide the features into two groups, corresponding to male and female. However, in the last layer, an interesting difference emerges: the tone-only model mixes the features by sex, while the tone and sex model continues to separate them. Despite this, both models perform well on the tone classification task. This observation suggests that there are multiple ways for a speech model to encode different tasks effectively.

\section{Conclusion}

 In conclusion, our study demonstrates that fine-tuning wav2vec 2.0 models effectively achieves phonetic normalization by selectively suppressing irrelevant information, such as speaker sex, while enhancing task-relevant features, like tones and finals. We found that models fine-tuned for multiple tasks retain information for both tasks without compromising performance and that suppressing task-irrelevant information is not necessary for effective classification. These findings provide new insights into how phonetic normalization can be flexibly achieved in speech models and how it is realized in human speech perception.
 

\bibliographystyle{IEEEbib}
\bibliography{mybib}

\end{document}